\documentclass[journal]{IEEEtran}
\usepackage{amsmath,amssymb,amsfonts}
\usepackage{tabularx,array,booktabs}
\usepackage{multirow}
\usepackage{algorithmic}
\usepackage{graphicx}
\usepackage{textcomp}
\usepackage{xcolor,colortbl}
\usepackage{epstopdf}
\usepackage{bm}
\usepackage{subfigure}
\usepackage{tikz}
\usetikzlibrary{shapes,arrows,fit,positioning}
\usepackage{cite}

\newcolumntype{C}{>{\centering\arraybackslash}m{2.8cm}}
\newcolumntype{A}{>{\centering\arraybackslash}m{1.19cm}}
\newcommand{\cellc}{\cellcolor{gray!20}}

\begin{document}

\title{Adversarial Deep Learning in EEG Biometrics}

\author{Ozan~\"{O}zdenizci, Ye~Wang, Toshiaki~Koike-Akino, and~Deniz~Erdo\u{g}mu\c{s}%
\thanks{O.~\"{O}zdenizci and D.~Erdo\u{g}mu\c{s} are with the Cognitive Systems Laboratory at Department of Electrical and Computer Engineering, Northeastern University, Boston, MA, USA. E-mail: \{oozdenizci, erdogmus\}@ece.neu.edu.}%
\thanks{Y.~Wang and T.~Koike-Akino are with the Mitsubishi Electric Research Laboratories, Cambridge, MA, USA. E-mail: \{yewang, koike\}@merl.com.}%
\thanks{O. \"{O}. and D. E. are partially supported by NSF (IIS-1149570, CNS-1544895), NIDLRR (90RE5017-02-01), and NIH (R01DC009834). Authors thank Paula Gonzalez-Navarro for her contributions in data collection.}%
}

\markboth{IEEE Signal Processing Letters}%
{\"{O}zdenizci \MakeLowercase{\textit{et al.}}: Adversarial Deep Learning in EEG Biometrics}

\maketitle

\begin{abstract}Deep learning methods for person identification based on electroencephalographic (EEG) brain activity encounters the problem of exploiting the temporally correlated structures or recording session specific variability within EEG. Furthermore, recent methods have mostly trained and evaluated based on single session EEG data. We address this problem from an invariant representation learning perspective. We propose an adversarial inference approach to extend such deep learning models to learn session-invariant person-discriminative representations that can provide robustness in terms of longitudinal usability. Using adversarial learning within a deep convolutional network, we empirically assess and show improvements with our approach based on longitudinally collected EEG data for person identification from half-second EEG epochs.\end{abstract}


\IEEEpeerreviewmaketitle

\section{Introduction}
\label{sec:intro}

Non-invasively recorded electroencephalographic (EEG) human brain activity has gained interest as an alternative person-discriminative biometric due to its continuous accessibility, privacy compliancy, and relatively harder forgeability, in comparison to today's most prevalent biometric identification approach of fingerprint recognition. A significant amount of work within the field of statistical EEG signal processing proposed novel methodologies to explicitly access person-discriminative neural sources from EEG. This problem was successfully tackled both in the context of \textit{person identification}, where an individual is assigned to a label (class) within a specific set of people that an identification model is trained on \cite{Campisi:2014}, as well as for \textit{person authentication}, where a one-to-one matching in decision making for person recognition is performed \cite{Marcel:2007}. Longitudinal studies also confirm the feasibility of EEG as an alternative means of biometrics \cite{Das:2016,Maiorana:2016,Maiorana:2018}. However, one recent study demonstrates different affective mental states potentially influencing stability of EEG as a tool for user identification \cite{Arnau:2018}. As such, discriminative EEG biometric feature extractor models that can filter out specific nuisance variables are likely to enhance usability of generated invariant features for biometric identification. Similarly, this idea can extend to filtering out recording session related confounders from the feature learning process for longitudinal model robustness.

Recent progress in EEG deep learning has capabilities to tackle this problem. However importantly, current deep learning models for EEG biometric identification are vastly evaluated by within-session (i.e., within-recording) cross-validation protocols \cite{Ma:2015,Mao:2017,Zhang:2018,Wilaiprasitporn:2018}. Due to their deep and complex nature, these models are particularly prone to capturing recording-specific variability rather than individual neural biomarkers. Going further, such models rely on the hypothesis that the deep architectures will internally learn invariant, generalizable features. This assumption is naturally constrained with the amount of available person-representative EEG data. Here, we highlight the need to extend EEG neural network models to explicitly learn invariant features against longitudinal variabilities.

In this study, we present an adversarial inference approach to extend deep learning based EEG biometric identification models, to learn session-invariant person-discriminative features (representations). We evaluated our approach based on EEG data recorded from ten healthy subjects participating in rapid serial visual presentation (RSVP) based brain-computer interface (BCI) experiments, which are identically performed on three separate sessions (i.e., days) for each participant. Empirical evaluations revealed a significant improvement with adversarial session-invariant feature learning for across-sessions person identification compared to conventional methods.

\section{Prior Work}
\label{sec:relatedwork}

\subsection{EEG-Based Biometric Identification}
\label{sec:eegbiometrics}

Pioneering neural signatures in EEG biometrics were considered to be visually evoked potentials (VEPs) \cite{Campisi:2014}. Several pieces of work investigated spatio-temporal dynamics of EEG during visual stimuli perception for person identification \cite{Das:2009,Armstrong:2015,Touyama:2008,Koike:2016,Palaniappan:2007}. Going further, since VEP-based experiment designs would require high physical and mental user attention, various studies extended this interest to different settings. Some examples include EEG recordings during silently reading texts \cite{Gui:2014}, thorough EEG time-frequency domain explorations during emotion elicitation, resting-state, or motor imagery/execution tasks \cite{DelPozo:2015}, fully task-independent designs where data are collected during various kind of auditory stimuli that does not require particular attention \cite{Vinothkumar:2018}, or imagined speech \cite{Brigham:2010}. An alternative approach is multitask learning in EEG biometrics, which was addressed in a study where person identification and motor task prediction was performed simultaneously through a shared representation to take advantage of latent task-specific informations \cite{Sun:2008}. Overall, these methods were mostly explored by data set specific traditional EEG processing methods that uses time-domain features \cite{Das:2009,Armstrong:2015,Touyama:2008,Koike:2016}, spectral decompositions \cite{Palaniappan:2007,Gui:2014,DelPozo:2015,Vinothkumar:2018} or autoregressive coefficients \cite{Brigham:2010}.

Motivated by its rapid progress, deep neural networks have recently gained significant interest as generic spatio-temporal EEG feature extractors. Mainly structured with convolutional architectures, deep neural networks were introduced for P300 detection \cite{Cecotti:2011}, steady-state visually evoked potential detection \cite{Kwak:2017}, rhythm perception during auditory stimuli \cite{Stober:2014}, decoding of motor imagery \cite{Sakhavi:2018}, as well as recently for non-task-specific discriminative EEG feature extraction \cite{Schirrmeister:2017,Lawhern:2018}. Such convolutional neural networks (CNNs) were also extended to recurrent-CNNs \cite{Bashivan:2016}, as well as deep convolutional autoencoders \cite{Stober:2016}. However, minimal progress was recently made in using these generic EEG feature extractors for biometrics. Some examples explore CNNs on resting-state \cite{Ma:2015} and motor imagery EEG \cite{Das:2018}, recurrent neural networks (RNNs) \cite{Zhang:2018}, as well as a combination of CNNs and RNNs for person-discriminative feature extraction under different mental states \cite{Wilaiprasitporn:2018}. Similarly, a recent task-independent approach applies deep networks to EEG data recorded during driving \cite{Mao:2017}.

Although resting-state EEG or recordings under different mental states \cite{Arnau:2018} may prove a baseline for task-independence, models that can filter out any underlying cued states to learn invariant features would be useful in EEG biometrics. Furthermore, given that deep learning models can easily capture recording-specific artifacts rather than individual EEG biomarkers, feature invariance across recording sessions would be of particular interest for longitudinal usability. To this end, existing studies rely on deep capabilities of the networks to learn invariant and robust biometric EEG features when a large pool of data is used. This assumption can be constrained with the amount of person-representative EEG data. Going further, recent works mostly evaluate their methods based on within-session model learning and testing \cite{Ma:2015,Mao:2017,Zhang:2018,Wilaiprasitporn:2018}. Hence, explicitly learning session-invariant biometric representations with deep learning remained as an open question for exploration.

\subsection{Adversarial Invariant Representation Learning}
\label{sec:adversariallearning}

Adversarial learning has been successfully applied in many deep learning applications to date, mainly popularized within generative model approaches for image data augmentation \cite{Goodfellow:2014}. Adversarial training within generative models (e.g., variational autoencoder (VAE)) were also used for invariant latent representation learning, to disentangle specific attributes (e.g., nuisance variables) from the representations \cite{Edwards:2016,Louizos:2016,Lample:2017}. These architectures rely on training a generative model objective (e.g., evidence lower bound for VAE \cite{Kingma:2014}), alongside a competing adversary with an antagonistic feedback on the overall optimization objective to enforce the invariance. Such attribute-invariant latent representations can then be used to manipulate these attributes in data augmentation \cite{Lample:2017}.

In our interest, there exists significant work on learning discriminative, attribute-invariant encoder models that do not require a generative decoder counterpart. In a discriminative context, features are learned through an encoder during an adversarial training game, by maximizing label prediction certainty from learned features, while minimizing the certainty of inferring the attribute (e.g., nuisance) variables from these features \cite{Xie:2017,Louppe:2017}. To date, these advancements in invariant feature learning have not been considered in EEG biometrics. In the light of these works, we hypothesize that adversarial discriminative inference can be useful in terms of session-invariant deep EEG feature learning for biometrics.

\section{Adversarial Convolutional Network}
\label{sec:methods}

\subsection{Adversarial Model Learning}
\label{sec:advinference}

Let $\{(\bm{X}_i,s_i,r_i)\}_{i=1}^{n}$ denote a model training data set, with $\bm{X}_i\in\mathbb{R}^{C \times T}$ the raw EEG data at \textit{epoch} $i$ recorded from $C$ channels for $T$ discretized time samples, $s_i \in \{1,\ldots,S\}$ the subject identification (ID) number for the person that the EEG is collected from, and $r_i \in \{1,2,\ldots,R\}$ the recording session ID (i.e., day) that the EEG data of the subject is collected at. EEG data generation process is assumed to be jointly dependent on subject and recording session IDs (i.e., $\bm{X} \sim p(\bm{X} \vert s,r)$). In our decoding problem, the class labels from the discriminative perspective are the subject IDs, and we aim to learn person identification models over a specified number of subjects $S$ in our training data set.

In the proposed adversarial discriminative model learning framework, we train a deterministic convolutional \textit{encoder} $g(\bm{X};\theta)$ with parameters $\theta$, that will ideally output representations which are predictive of $s$ as recovered by an \textit{identifier} network parameterized by $\gamma$, but not predictive of $r$ as concealed from an \textit{adversary} network parameterized by $\phi$ which tries to recover $r$. Here, the parametric identifier models the likelihood $q_{\gamma}(s\vert g(\bm{X};\theta))$, whereas the parametric adversary models the likelihood $q_{\phi}(r\vert g(\bm{X};\theta))$. While the adversary is trained to maximize $q_{\phi}(r\vert g(\bm{X};\theta))$, the encoder conceals information regarding $r$ from the learned representations by minimizing this likelihood, as well as also retaining person-discriminative information by maximizing $q_{\gamma}(s\vert g(\bm{X};\theta))$. This results in jointly training the networks towards the objective:
\begin{equation}
\min_{\theta,\gamma} \max_{\phi} \mathbb{E}[-\log q_{\gamma}(s\vert g(\bm{X};\theta)) + \lambda\log q_{\phi}(r\vert g(\bm{X};\theta))],
\label{eq:objective}
\end{equation}
with $\lambda>0$ denoting the adversarial loss weight to enforce stronger invariance and trading-off with identification performance. Optimization is performed by using stochastic gradient descent alternatingly for the adversary, and the encoder-identifier networks to optimize Eq.~\ref{eq:objective}. An overview of the adversarial training framework is illustrated in Figure~\ref{fig:architecture}.

\subsection{Convolutional Network Architecture}
\label{sec:cnnarchitecture}

Our EEG feature encoder consists of four convolutional blocks (c.f.~Table~\ref{tab:network}), mainly structured by field-leading works \cite{Schirrmeister:2017,Lawhern:2018}. Sequentially, we perform temporal convolutions resembling to frequency filtering, depthwise convolutions \cite{Chollet:2017} as spatial filtering of frequency-specific activity, and two more 2D convolution blocks for spatio-temporal feature aggregation. After convolutions we use batch normalizations (BatchNorm) \cite{Ioffe:2015}, and rectified linear unit (ReLU) activations. We use ReLU activations since they typically learn faster in networks with many layers \cite{LeCun:2015}, and were successfully used in EEG deep learning \cite{Bashivan:2016}. We did not observe performance increases by using exponential linear units \cite{Lawhern:2018}. Deep convolution kernel sizes are mostly guided by \cite{Schirrmeister:2017}, since we observed training set overfitting and overall lack of generalization with shallower layers. Beyond architectural design choices, in fact, our adversarial inference approach is applicable to any EEG neural network model by simply modifying the training objective. 

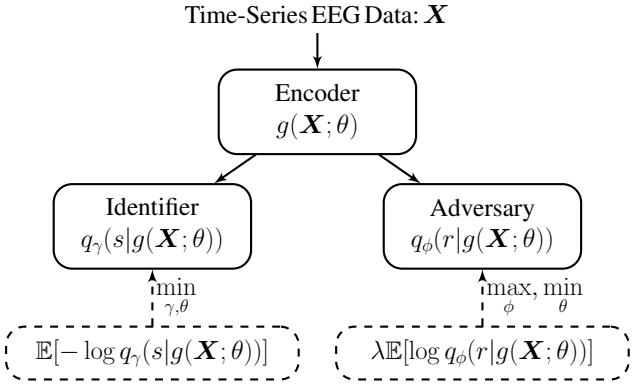
\begin{figure}[t]
	\centering
	\begin{tikzpicture}[thick,scale=0.4, every node/.style={transform shape}]
    \tikzstyle{empty} = [cloud, circle, text width=6em, text centered, minimum height=2em]
    \tikzstyle{data} = [cloud, rectangle, rounded corners=7pt, fill=white!10, text width=40em, minimum height=3.5em, text centered]
    \tikzstyle{block} = [cloud, rectangle, rounded corners=7pt, draw, fill=white!10, text width=18em, minimum height=8em, text centered]
    \tikzstyle{loss} = [cloud,  rectangle, rounded corners=7pt, dashed, draw, fill=white!10, text width=27em, minimum height=5em, text centered]
    \tikzstyle{arrow} = [draw, -latex']

    \node [data] (EEG) {\Huge Time-Series EEG Data: $\bm{X}$};
    \node [block, below = 1.2cm of EEG] (Enc) {\Huge Encoder\\\vspace{0.4cm}$g(\bm{X};\theta)$};
    \node [empty, below = 1.2cm of Enc] (BOS) {};
    \node [block, left = 1cm of BOS] (Ide) {\Huge Identifier\\\vspace{0.4cm}$q_\gamma(s\vert g(\bm{X};\theta))$};
    \node [block, right = 1cm of BOS] (Adv) {\Huge Adversary\\\vspace{0.2cm}$q_\phi(r\vert g(\bm{X};\theta))$};
    \node [loss, below = 1.9cm of Adv] (Ladv) {\Huge $\lambda\mathbb{E}[\log q_{\phi}(r \vert g(\bm{X};\theta))]$};
    \node [loss, below = 1.9cm of Ide] (Lide) {\Huge $\mathbb{E}[-\log q_{\gamma}(s \vert g(\bm{X};\theta))]$};
    
    \draw [arrow,dashed] (Lide) -- node [right] {\Huge$\displaystyle\min_{\gamma,\theta}$} (Ide);
    \draw [arrow,dashed] (Ladv) -- node [right] {\Huge$\displaystyle\max_{\phi},\min_{\theta}$} (Adv);
    \draw [arrow] (EEG) -- (Enc);
    \draw [arrow] (Enc) -- (Ide);
    \draw [arrow] (Enc) -- (Adv);
\end{tikzpicture}
	\caption{Adversarial discriminative model training framework. Encoder, identifier and adversary networks are simultaneously trained towards the objective in Eq.~\ref{eq:objective}, as illustrated by the loss functions in the dashed boxes.}
	\label{fig:architecture}
\end{figure}

Representations learned by the encoder were used by the identifier and adversary for linear classifications of subject and session IDs. Both networks consisted of a fully-connected layer with $S$ or $R$ softmax units, respectively for the identifier and adversary, to obtain the normalized log-probabilities represented in the loss functions as shown in Figure~\ref{fig:architecture}.

\section{Experimental Study}
\label{sec:results}

\subsection{Study Design and Experimental Data}
\label{sec:daq}

Ten healthy subjects participated in the experiments at three identical sessions performed on different days. The average interval between two consecutive recording sessions for the same person was 7.85 $\pm$ 14.34 days, with a minimum of 1 and a maximum of 65 days between sessions across all experiments. Before the experiments, all participants gave their informed consent in accordance with the guidelines set by the research ethics committee of Northeastern University. 

During each session, EEG data were recorded from participants while they were using the RSVP Keyboard\texttrademark, an EEG-based BCI speller that relies on the RSVP paradigm to visually evoke event-related brain responses for user intent detection \cite{Orhan:2012}. EEG data were recorded from 16 channels (as described and also used in previous work \cite{Gonzalez:2016,Gonzalez:2017}), sampled at 256 Hz, using active electrodes and a g.USBamp biosignal amplifier (g.tec medical engineering GmbH, Austria).

Participants were using the RSVP Keyboard\texttrademark~in offline calibration mode, attending to pre-specified letters while random sequences of visual letter stimuli are presented to them. EEG data were epoched at [0-0.5] seconds post-stimuli intervals to construct each subjects' specific session data set. We pooled all epochs within a recording session irregardless of their attended versus non-attended stimuli labels. The complete data set consisted of 41,400 epochs of 16 channel EEG data for 128 samples. Epochs were equally distributed across subjects and sessions (i.e., 1,380 epochs per subject per session).

\renewcommand{\arraystretch}{1.1}
\begin{table}
	\caption{Convolutional Encoder Network Specifications}
    \vspace{-0.1cm}
	\label{tab:network}
	\begin{center}
	\begin{tabular}{c|c|c}
		\toprule
		\textbf{Layer} & \textbf{Operation} & \textbf{Output Dim.} \\
		\midrule
		Encoder Input & Reshape & $(1,C,T)$ \\
		\midrule
		\multirow{2}{*}{\parbox{2cm}{\centering Convolutional Block 1}} & $20 \times$ Conv $(1,T/2)$ & $(20,C,T/2)$ \\
		& BatchNorm & $(20,C,T/2)$ \\
		\midrule
		\multirow{2}{*}{\parbox{2cm}{\centering Convolutional Block 2}} & $20 \times$ DepthwiseConv $(C,1)$ & $(400,1,T/2)$ \\
		& BatchNorm + ReLU + Reshape & $(1,400,T/2)$ \\
		\midrule
		\multirow{2}{*}{\parbox{2cm}{\centering Convolutional Block 3}} 
		& $200 \times$ Conv $(400,T/4)$ & $(200,1,T/4)$ \\
		& BatchNorm + ReLU + Reshape & $(1,200,T/4)$ \\
		\midrule
		\multirow{2}{*}{\parbox{2cm}{\centering Convolutional Block 4}} 
		& $100 \times$ Conv $(200,T/8)$ & $(100,1,T/8)$ \\
		& BatchNorm + ReLU & $(100,1,T/8)$ \\
		\midrule
		Encoder Output & Flatten & $100*T/8$ \\
	    \bottomrule
	\end{tabular}
	\end{center}
    \vspace{-0.1cm}
\end{table}

\renewcommand{\arraystretch}{1.14}
\begin{table*}
  \caption{Leave-one-session-out person identification accuracies ($\%$). Test columns show the identifier accuracies for the left-out session. First two rows denote the baseline methods. Non-Adversarial model denotes a regular deep CNN. Adversarial models learn invariant features across two training sessions. Parentheses denote the standard deviations across 10 repetitions.}
  \label{tab:rsvp}
  \centering
  \begin{tabular}{C A|A|A|A|A|A|A|A|A}
    \toprule
    & \multicolumn{3}{c|}{\textbf{Learning on Session 2 and 3}} & \multicolumn{3}{c|}{\textbf{Learning on Session 1 and 3}} & \multicolumn{3}{c}{\textbf{Learning on Session 1 and 2}} \\
    \cmidrule{2-10}
    & \multicolumn{2}{c|}{\textbf{Validation Set}} & \multirow{2}{*}{\parbox{1.2cm}{\centering \textbf{Test Session 1}}} & \multicolumn{2}{c|}{\textbf{Validation Set}} & \multirow{2}{*}{\parbox{1.2cm}{\centering \textbf{Test Session 2}}} & \multicolumn{2}{c|}{\textbf{Validation Set}} & \multirow{2}{*}{\parbox{1.2cm}{\centering \textbf{Test Session 3}}} \\
    \cmidrule{2-3} \cmidrule{5-6} \cmidrule{8-9}
    & \textbf{Identifier} & \textbf{Adversary} & & \textbf{Identifier} & \textbf{Adversary} & & \textbf{Identifier} & \textbf{Adversary} &  \\
    \midrule
    \multicolumn{1}{C|}{Spectral Powers + QDA}
    & $80.5$ \tiny{(.005)} & -- & \cellc $44.9$ \tiny{(.001)}
    & $81.4$ \tiny{(.004)} & -- & \cellc $52.1$ \tiny{(.001)} 
    & $83.6$ \tiny{(.004)} & -- & \cellc $49.5$ \tiny{(.002)} \\
    \multicolumn{1}{C|}{PCA + QDA}
    & $85.6$ \tiny{(.006)} & -- & \cellc $57.6$ \tiny{(.002)}
    & $85.1$ \tiny{(.006)} & -- & \cellc $58.9$ \tiny{(.001)} 
    & $88.1$ \tiny{(.002)} & -- & \cellc $64.1$ \tiny{(.002)} \\
    \multicolumn{1}{C|}{Non-Adversarial $\lambda=0$}
    & $91.6$ \tiny{(.008)} & $76.0$ \tiny{(.02)} & \cellc $62.3$ \tiny{(.02)} 
    & $97.9$ \tiny{(.006)} & $78.9$ \tiny{(.03)} & \cellc $63.2$ \tiny{(.02)} 
    & $97.9$ \tiny{(.001)} & $76.9$ \tiny{(.05)} & \cellc $69.2$ \tiny{(.02)} \\
    \multicolumn{1}{C|}{Adversarial $\lambda=0.005$}
    & $91.2$ \tiny{(.01)} & $65.2$ \tiny{(.01)} & \cellc $65.1$ \tiny{(.01)} 
    & $98.4$ \tiny{(.005)} & $63.7$ \tiny{(.03)} & \cellc $69.1$ \tiny{(.01)} 
    & $98.2$ \tiny{(.006)} & $60.4$ \tiny{(.03)} & \cellc $\mathbf{71.6}$ \tiny{\textbf{(.02)}} \\
    \multicolumn{1}{C|}{Adversarial $\lambda=0.01$} 
    & $90.7$ \tiny{(.01)} & $60.8$ \tiny{(.02)} & \cellc $\mathbf{66.6}$ \tiny{\textbf{(.02)}}
    & $98.4$ \tiny{(.002)} & $58.7$ \tiny{(.02)} & \cellc $\mathbf{69.2}$ \tiny{\textbf{(.02)}} 
    & $98.4$ \tiny{(.003)} & $58.5$ \tiny{(.03)} & \cellc $71.3$ \tiny{(.01)} \\
    \multicolumn{1}{C|}{Adversarial $\lambda=0.02$} 
    & $91.1$ \tiny{(.01)} & $57.2$ \tiny{(.04)} & \cellc $65.3$ \tiny{(.01)}
    & $98.1$ \tiny{(.003)} & $56.0$ \tiny{(.03)} & \cellc $68.7$ \tiny{(.03)}
    & $98.2$ \tiny{(.003)} & $54.9$ \tiny{(.04)} & \cellc $71.2$ \tiny{(.01)} \\
    \multicolumn{1}{C|}{Adversarial $\lambda=0.05$}
    & $91.0$ \tiny{(.009)} & $53.4$ \tiny{(.05)} & \cellc $65.4$ \tiny{(.02)}
    & $97.8$ \tiny{(.005)} & $53.5$ \tiny{(.03)} & \cellc $67.6$ \tiny{(.02)}
    & $98.2$ \tiny{(.003)} & $54.0$ \tiny{(.03)} & \cellc $71.1$ \tiny{(.02)} \\
    \multicolumn{1}{C|}{Adversarial $\lambda=0.2$}
    & $91.8$ \tiny{(.005)} & $54.5$ \tiny{(.03)} & \cellc $64.1$ \tiny{(.02)}
    & $97.1$ \tiny{(.006)} & $53.1$ \tiny{(.04)} & \cellc $66.4$ \tiny{(.02)}
    & $97.4$ \tiny{(.004)} & $53.5$ \tiny{(.03)} & \cellc $71.1$ \tiny{(.03)} \\
    \bottomrule
  \end{tabular}
  \vspace{-0.1cm}
\end{table*}

\subsection{Data Analysis and Implementation}
\label{sec:analysis}

We performed both within-session person identification analyses to illustrate conventional evaluation methods, as well as across-sessions analyses to demonstrate the impact of session-invariant feature learning. In within-session analyses, all subjects' data were pooled per session ID and three distinct within-session models were trained. \textit{Training}, \textit{validation} and \textit{test} data sets were randomly constructed as 70\%, 10\% and 20\% portions of the within-session data pools. Here, we ignore the adversary network and train the encoder-identifier as a regular CNN. Across-sessions analyses were performed by a leave-one-session-out approach, where the left-out session constituted the \textit{test} set, and the \textit{training} and \textit{validation} sets were constructed as 80\% and 20\% random splits of the other two sessions' pooled data. Here, the adversary was jointly trained to recover session IDs from the encoder outputs.

We further compare the deep CNN-based models with two baseline methods. First approach uses power spectrum features \cite{Palaniappan:2007,Gui:2014,DelPozo:2015,Vinothkumar:2018}. We concatenated channel log-bandpowers computed in $\theta$- (4-8 Hz), $\alpha$- (8-15 Hz), $\beta$- (15-20 Hz, 20-25 Hz, 25-30 Hz) and $\gamma$- (30-45 Hz, 45-60 Hz, 60-75 Hz) bands through FFT using the Welch's method and Hanning windows, into a feature vector. We used a quadratic discriminant analysis (QDA) classifier as the identifier, since it performed better than linear classifiers in our analyses. Second method uses principal component analysis (PCA) projection on the 2048 dimensional (16 times 128) vectorized EEG data, and a QDA classifier \cite{Touyama:2008,Koike:2016}. PCA projection dimensionality was determined as the minimum number of components that accounted for 90\% of total variance, which varied between 168 and 174.

All EEG data were normalized to have zero mean and scaled to [-1,1] range by dividing with the absolute maximum value at each epoch and channel individually. No channel selection or offline artifact removal was performed. Input dimensionality of the CNN networks is $C$=16 channels by $T$=128 samples. The identifier network has a 10-dimensional output as a $S$=10 class classifier across the subjects, whereas the adversary is a binary classifier across $R$=2 sessions. It is important to note that the binary session ID is not a shared variable across subjects, but simply indicates variability with its label in conjunction with the subject IDs. Ideally, alternating the session ID labels should not have a significant impact on the learned models.

Networks were trained with 100 epochs per batch, with 500 repeated training set passes. Early stopping was performed based on validation set loss of the identifier network. Parameters were updated once per batch with Adam \cite{Kingma:2015}. We used the Chainer deep learning framework for implementations \cite{Tokui:2015}.

\subsection{Within-Session Person Identification}
\label{sec:withinsession}

Models were learned using the training and validation splits of single session pooled data. Our CNN model was able to discriminate 10 subjects with 98.7\%$\pm$0.005, 99.3\%$\pm$0.003, and 98.6\%$\pm$0.006 accuracies for Sessions 1, 2, and 3 respectively, over 2,760 half-second epochs (i.e., 20\% test split).

\begin{figure}
	\centering
	\includegraphics[trim=0cm 0.3cm 1cm 0cm, width=0.48\textwidth]{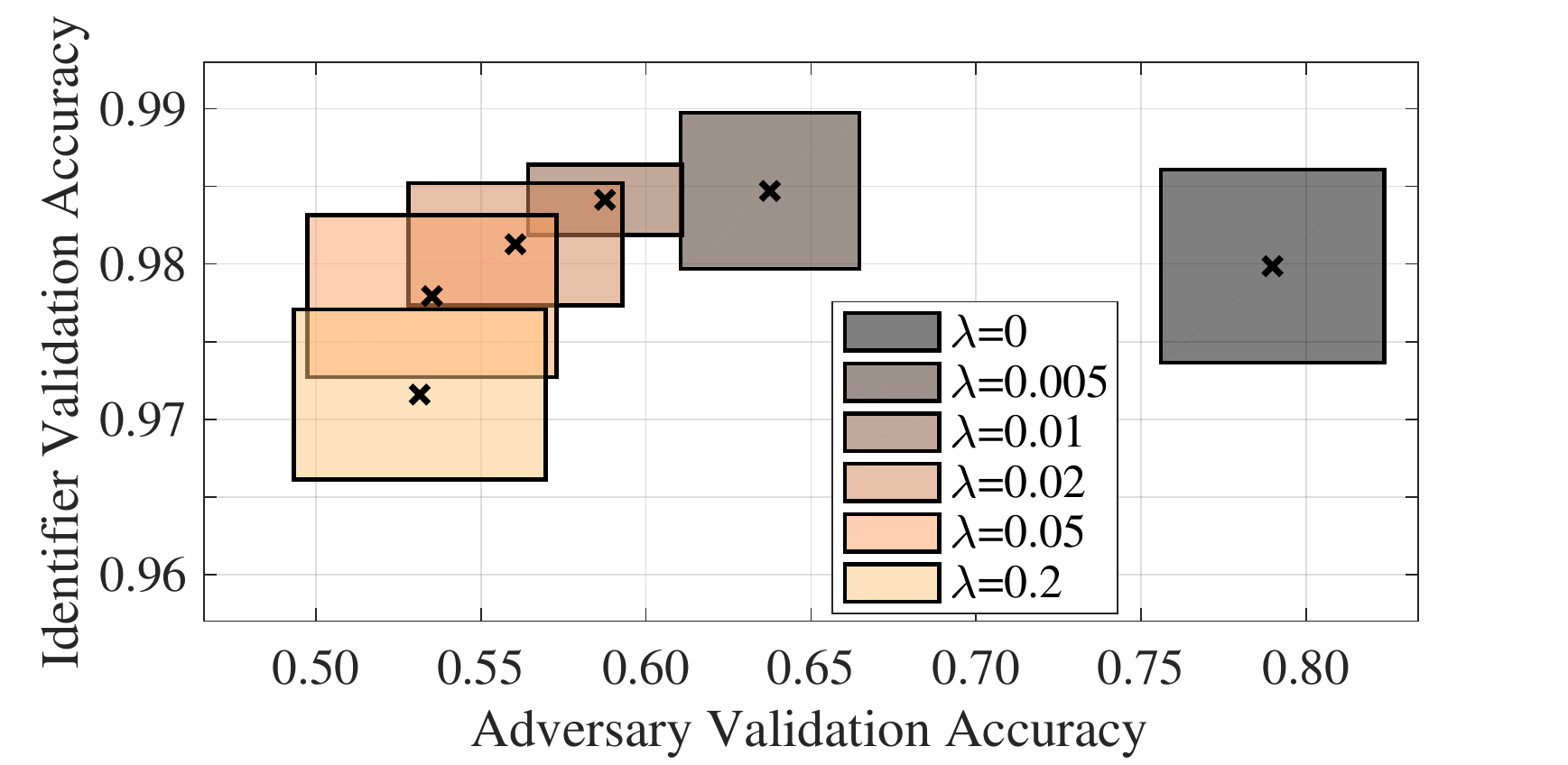}
	\caption{Adversarial ($\lambda>0$) and non-adversarial ($\lambda=0$) model evaluations with identifier and adversary validation accuracies for the leave-Session 2-out learning case. Center marks denote the means across ten repetitions and widths denote $\pm$1 standard deviation intervals in both dimensions.}
	\label{fig:modeltrain}
  \vspace{-0.1cm}
\end{figure}

\subsection{Across-Sessions Person Identification}
\label{sec:acrosssessions}

Since within-session learning and evaluation from EEG can overperform due to its highly correlated temporal structure, across-sessions analyses would better demonstrate model generalizability. Spectral power and time-domain representation based methods present a baseline with reasonable validation set performances, which do not generalize across-sessions well (c.f.~Table~\ref{tab:rsvp}). In non-adversarial models, we observe the amount of exploited session-discriminative leakage through the adversary we train alongside the CNN, without adversarial loss feedback ($\lambda=0$). We observed that regular CNNs exploit features that can also discriminate the two days (75--80\%).

Adversarial models suppress session-variant information from the encoded features. As observed from the adversary accuracies, increasing $\lambda$ censors the encoder, enforces stronger session-invariance, and converges to the 50\% chance level. An intuitive way of choosing $\lambda$ is by cross-validating the learning process. We train our models with varying $\lambda$, and favor decreases in adversary performance on the validation set with increasing $\lambda$, while maintaining a similar identifier performance compared to the non-adversarial case. Figure~\ref{fig:modeltrain} depicts (for the leave-Session 2-out case) that strong $\lambda$ values can force the encoder to lose person-discriminative information. Hence, we can choose $\lambda$ in a range where identifier does not start to perform poorly and adversary accuracy is low (e.g., $\lambda=0.01$ or $0.02$). When tested on an independent session, we observe up to 72\% 10-class person identification accuracies based on 13,800 half-second epochs, with up to $6\%$ gains via adversarial learning based on two sessions' invariance.

\section{Discussion}
\label{sec:discussion}

We propose an adversarial inference approach to extend deep learning based EEG biometric identification models, to learn session-invariant person-discriminative representations. We empirically assessed our approach based on half-second EEG epochs recorded from ten subjects during BCI experiments on three different sessions. Our results demonstrate significant contribution of adversarial learning in developing across-days EEG-based person identification models. 

Within-recording deep model learning and evaluation protocols are expected to perform significantly better when temporally correlated signals (e.g., EEG) are considered. Recently expanding work in EEG biometrics are mainly evaluated by these frameworks \cite{Ma:2015,Mao:2017,Zhang:2018,Wilaiprasitporn:2018}. Yet, one recent study that evaluates CNNs in longitudinal usability yielded significant insights to this problem \cite{Das:2018}. We address this in a similar way, while introducing adversarial learning for deep person-discriminative models to exploit session-invariant features. Overall, our approach is applicable to any EEG neural network model.

Our model currently relies on the assumption that changing sessions and days is one specific source of variability in the data distribution. This idea could well extend to task-invariant feature learning, where subjects can ideally perform any unspecified task that representations should be invariant to (e.g., naturalistic physical movements \cite{Nakamura:2018}). An ideal EEG person identification model would have no possibility to be calibrated or finetuned prior to use at an arbitrary time. Hence, in the light of recent progress in deep learning, we propose adversarial inference for longitudinal model robustness.



\end{document}